\documentclass[runningheads]{llncs}

\usepackage[utf8]{inputenc}
\usepackage{graphicx}
\usepackage{dirtytalk}
\usepackage{multirow}
\usepackage{tikz}
\usepackage{tabularx}
\usepackage{adjustbox}
\usepackage{enumitem}
\usepackage{makecell}
\usepackage{subcaption}
\usepackage{caption} 
\usepackage{hyperref}
\usepackage{tabu}
\usepackage{caption}

\newcommand{\itelos}{\textit{iTelos}}

\title{Building Interoperable Electronic Health Records as Purpose Driven Knowledge Graphs\thanks{This paper is a major extension of the short paper \cite{KD-2022-Giunchiglia-Popularity}. Sections 1,2,6,7 are completely new while Sections 3,4,5 are major extensions of the work described in \cite{KD-2022-Giunchiglia-Popularity}, largely focused of the application of \itelos\ in the Health domain.}}

\author{
Simone Bocca \and Alessio Zamboni \and Gábor Bella \and Yamini Chandrashekar \and Mayukh Bagchi \and Gabriel Kuper \and Paolo Bouquet \and Fausto Giunchiglia}

\institute{Department of Information Engineering and Computer Science (DISI),\\ University of Trento, Italy\\
}

\date{March 2021}

\begin{document}

\maketitle

\begin{abstract}
\vspace{-0.2cm}

When building a new application we are increasingly confronted with the need of reusing and integrating pre-existing knowledge. Nevertheless, it is a fact that this prior knowledge is virtually impossible to reuse \textit{as-is}. This is true also in domains, e.g., \textit{eHealth}, where a lot of effort has been put into developing high-quality standards and reference ontologies, e.g. \texttt{FHIR}\footnote{\url{https://hl7.org/fhir/}.}. In this paper, we propose an integrated methodology, called \itelos, which enables data and knowledge reuse towards the construction of \textit{Interoperable Electronic Health Records} (\texttt{iEHR)}. 
The key intuition is that the \textit{data level} and the \textit{schema level} of an application should be developed independently,  thus allowing for maximum flexibility in the reuse of the prior knowledge, but under the overall guidance of the needs to be satisfied, formalized as \textit{competence queries}. This intuition is implemented by codifying all the requirements, including those concerning \textit{reuse}, as part of a \textit{purpose} defined \emph{a priori}, which is then used to drive a \textit{middle-out} development process where the application schema and data are continuously aligned. The proposed methodology is validated through its application to a large-scale case study.
\vspace{-0.2cm}

\end{abstract}


\section{Introduction} \label{sec1_introduction}

Once upon a time, one would design an application \textit{top-down} starting from the requirements down to implementation, without thinking of the data: they would be generated by the system in production with, at most, the need of initializing them with data from the legacy systems being substituted. Nowadays, more and more, we are designing systems which, at the beginning but also when in production, must be integrated with data coming from other systems, possibly from third parties. Some examples are health systems that integrate personal data coming from multiple institutions and B2C (Business-to-Consumer) applications exploiting the big data available on the Web, e.g., open data or streaming data. 

The key aspect of this reuse problem is how to handle the \textit{semantic heterogeneity} which arise any time there is the need to perform data integration across multiple sources \cite{2021-KGCW}. This problem has been extensively studied in the past and two main approaches have been proposed. 
The first is using \textit{ontologies} to agree on a fixed language or schema to be shared across applications \cite{2017-obdi-for-mdee}. 
The second is the use of \textit{Knowledge Graphs (KGs)} and the exploitation of the intrinsic  flexibility and extensibility they provide \cite{2019-Kejriwal-KG}, as the means for facilitating the adaptation and integration of pre-existing heterogeneous data.
However, the problem is still largely unsolved. When developing an application, no matter whether one exploits ontologies or KGs, it is impossible to reuse the pre-existing knowledge \textit{as-is}. There is always some specificity which makes the current problem in need of dedicated development, with the further drawback that the resulting application is, again, hardly reusable. 
\\ \indent We propose  a general purpose methodology, called \itelos, whose main goal is to minimize as much as possible the negative effects of the above phenomenon.
 \itelos\ exploits all the previous results, in particular, it is crucially based on the use of ontologies and KGs. At the same time, \itelos\ takes a step ahead by providing a precise specification of the process by which an application should be developed, focusing on how to effectively reuse data from multiple sources. \itelos\ is based on three key assumptions:
 
 \begin{itemize}
 
 \item the \textit{data level} and the \textit{schema level} of an application should be developed independently,  thus allowing for maximum flexibility in the reuse of the prior data and schemas, e.g., ontologies, but under the overall guidance of  the needs to be satisfied, formalized as \textit{competence queries};
 \item Data and schemas to be reused, as well as competence queries, should be decided before starting the development, as precisely as possible, and defined \emph{a priori} as part of an application \textit{purpose}. Additionally, the purpose is assumed to specify a set of constraints specifying how much the satisfaction of each of its three elements is allowed to influence the satisfaction of the other two;
 \item the purpose should be used to drive a \textit{middle-out} development process where the successive evolutions of the application schema and data, both modeled as KGs, are continuously aligned \textit{upwards} (bottom-up) with the reference schemas to be reused and \textit{downwards} (top-down) with the data to be reused.
 \end{itemize}
The three assumptions listed above, which are at the core of the \itelos\ methodology, fit well with applications which arise in the \textit{eHealth} domain and in particular in the generation, integration and adaptation of \textit{Interoperable Electronic Health Records} (\texttt{iEHR}s for short). First of all, the clear separation between data and schemas is intrinsic to all health applications for at least two reasons. The first is the richness of standards which exist both at the schema level and at the data level (see, e.g., \cite{2020-Bella3,KD-2022-Bocca.a}). Keeping them separated allows for the incremental alignment of the KG modeling an \texttt{iEHR}. The second is that it allows for the incremental integration, first of the schemas and then of the data, the latter being much more critical being, differently from the schemas, highly sensible data, subject to very strict privacy rules. 
Moving to the second assumption listed above, the need for correctness and precision in the \textit{eHealth} domain, is supported by \itelos\ through the definition of the application \textit{purpose}. The definition of the purpose allows to define precisely which data and schema resources have to be considered for the production of a KG able to represent, and exploit an \texttt{iEHR}. Finally, the third assumption simply provides the guidelines for implementing the incremental generation of \texttt{iEHR}s, mentioned above, enabled by the separation between data and schemas.

The main goal of this paper is to report the lessons learned in the application of the \itelos\ methodology in the Health domain and in particular in the generation of \texttt{iEHR}s, as part of the European project \texttt{InteroEHRate}\footnote{The details of the project can be found at the URL \url{https://www.interopehrate.eu/}.}.
This paper is organized as follows. 
In Section \ref{sec2-purpose} we provide a description of the purpose and of how it is organized. 
In Section \ref{sec4_process} we provide a highlight of the \itelos\ middle-out process. In Section \ref{sec5_reuse}  we provide a detailed description of how \itelos\ enables the reusability of the available data. In Section
\ref{sec6_share} we describe how \itelos\ enables the sharability and future reusability of the application KG. In Section \ref{sec7_evaluation} we provide the main alignment metrics used to ensure that the \itelos\ middle-out process stays within the constraints specified by the purpose. Section \ref{sec8_case} describes how \itelos\ has been adopted in the \texttt{InteropEHRate} EU project. Finally, Section \ref{sec9_conclusion} closes the paper with the conclusions.

\nocite{1995reasoning}
\begin{figure}[ht!]
    \vspace{-0.6cm}
    \centering
    \includegraphics[width=8cm,height=4cm]{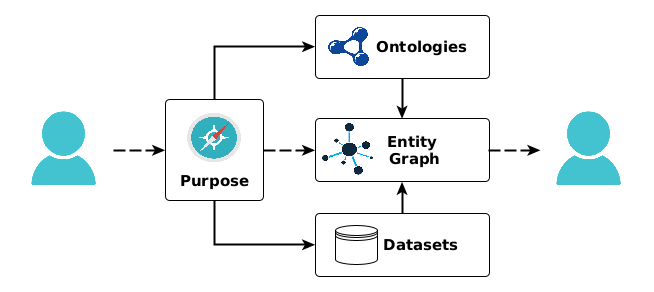}
    \vspace{-0.2cm}
    \caption{The \itelos\ approach.}
    \label{fig:approach}
    \vspace{-0.3cm}
\end{figure}

\section{The purpose}
\label{sec2-purpose}

The \itelos\ process is depicted at a very high level in Figure \ref{fig:approach}. The logical sequence, represented by the dashed lines, shows the \textit{User} providing in input an informal specification of the problem she wants to solve, the \textit{Purpose}, while receiving in output a KG, named, in Figure \ref{fig:approach}, the \textit{Entity Graph}(EG). The concrete process is represented by the four solid lines, indicating how the purpose leads to the reuse of prior knowledge, represented at the data level as \textit{Datasets} and at the schema level as \textit{Ontologies}, with the aim of building the Entity Graph.

The purpose, as the main input of the process, is composed of three elements:

\begin{itemize}
\vspace{-0.1cm}
    \item the functional requirements of the final application, in other words, the information that the final EG must be able to provide to satisfy the user's need. In the concrete example of the IEHR project, these requirements list the medical data to be included in an iEHR. We assume that such needs are formalized as a list of \textit{competency queries (CQ)} \cite{gruninger1995role}. 
    \item a set of datasets to be reused as already existing knowledge, thus, in turn, to be integrated into the final EG. A key assumption, as part of the overall \itelos\ strategy, is to handle such datasets as KGs, thus facilitating their future reuse. In the IEHR project, four different hospital partners provided the datasets to be considered for the purpose.
    \item a set of existing well-known \textit{reference schemas}, i.e., ontologies, but not only, to be reused in order to develop the EG's purpose-specific schema, which can then be shared for future applications. It is important to notice that well-known schemas are already available; for instance, LOV, LOV4IoT, and DATAHUB,\footnote{See, respectively: \url{https://lov.linkeddata.es/}, \url{http://lov4iot.appspot.com/}, \url{https://old.datahub.io/}.} are three among the most relevant repositories. However, for the IEHR project, the reference schemas of the datasets provided by hospitals have been mainly considered with a strong adoption of the FHIR reference ontology. In line with the reuse approach, to support the \itelos\ process, a new repository, called \textit{LiveSchema},\footnote{See: \url{http://liveschema.eu/}.} is under construction, where reference schemas are annotated by a very rich set of metadata, see, e.g., \cite{dutta2015mod,2020-KR}, with the goal of automating as much as possible the \itelos\ process.
\vspace{-0.1cm}
\end{itemize}
A crucial design decision in the structure of the purpose, which reflects the \itelos\ process structure depicted in Figure \ref{fig:approach}, is the separation of the data and schema level, by keeping them distinct and independent, as well as modelled as two different types of KG. This major aspect allows splitting the problem of reusing the existing datasets during the integration process from the problem of developing a unique schema for the final EG which can be easily reused in future applications.
%
The KGs considered for the data level, that we call \textit{Entity Graphs (EGs)}, are graphs having \textit{entities} as nodes (e.g., my dog \textit{Pluto}). The entities are composed of data property values used to describe them. The links in the data layer KGs are object properties describing the relations between any two entities. The schema level KGs, instead, are called \textit{Entity Type (etype) Graphs (ETGs)}, namely graphs defining the schema of an EG. Therefore, for each EG there is a corresponding ETG which defines its schema. The nodes of an ETG are \textit{etypes}, namely classes of entities (e.g., the class \textit{dog)}, Each etype is described (and actually, represented) by a set of data properties and by a set of object properties, defining the schema of each single etype and the possible relations among them, respectively.

\textit{Datasets} and \textit{Ontologies} depicted in Figure \ref{fig:approach} are examples of EGs and ETGs, respectively.
\itelos\ use EGs to represent: (i) the entity graph produced as the outcome of the whole process (see Figure \ref{fig:approach}) and, if they are available in this specific form, (ii) the input datasets. The ETGs, instead are exploited to represent: (i) the schema of the final entity graph, (ii) the input reference schemas to be reused, as well as, after a formalization process, (iii) the functional requirements extracted from the initial competency queries. \itelos\ maintains this uniformity of representation in order to exploit, for both data and schema layer, the high capacity of KGs to be composed of each other, as well as to be extended and adapted to different purposes. In line with the underlying approach, these KG features allow \itelos\ to reuse not only existing knowledge but also produce reusable and interoperable data which, in turn, reduces the effort in building future applications.


\begin{figure}[ht!]
    \centering
    \makebox[\textwidth][c]{\includegraphics[width=15cm,height=3.8cm]{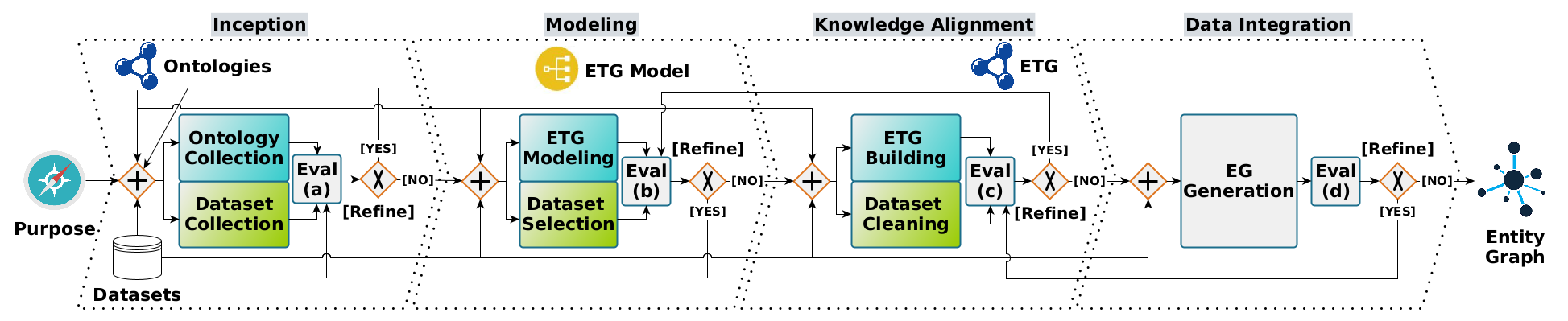}}
    \caption{The \itelos\ Process.}
    \label{fig:pipeline}
\end{figure}

\section{\itelos\ as a process} \label{sec4_process}

Figure \ref{fig:pipeline}, freely adapted from \cite{KD-2022-Giunchiglia-Popularity}, describes the key logical phases of the \itelos\ process (dashed line in Figure \ref{fig:approach}). 
The process depicted in  Figure \ref{fig:pipeline} is articulated in four phases generating an Entity Graph starting from the initial Purpose. The objective of each phase can be synthetically described as follows:

\begin{itemize}

    \item \textit{Inception}: the purpose, provided in input by the user, is formalized into a set of CQs, as well as used to identify and collect a set of existing, reusable, datasets and ontologies; 
    \item \textit{Modeling}: a purpose-specific model of the ETG is built by taking into account CQs and datasets;
    \item \textit{Knowledge Alignment}: a reusable ETG is built, based on the model designed previously, by reusing the selected reference ontologies;
    \item \textit{Data Integration}: the final EG is built by integrating the input datasets into the ETG.

\end{itemize}
Let us analyze these phases in detail. 
The \textit{Inception} phase takes the purpose from the user, initially specified as a natural language description of the desired objective. The functional requirements are extracted from the purpose and formalized into a list of CQs. In the \textit{Ontology Collection} and \textit{Dataset Collection} activities, shown in Figure \ref{fig:pipeline}, the reusable datasets and ontologies, being part of the purpose, are matched with the CQs in order to select the most suitable resources to build up the final EG. The final collection of data and schema resources may be extended by resources initially not considered in the list provided by the initial purpose. The key observation is that matching CQs and (the schemas of the) datasets is crucial for the success of the project. A little coverage would mean that there are not enough data for the implementation of the EG, therefore a revision of the CQs, or a more refined data collection is required. If the former is the case the process we follow is inspired by the work in \cite{gruninger2019ontology}.


The \textit{Modeling} phase receives in input the ontologies and datasets previously collected, as well as the CQ list. The main objective of this phase is to extract from the CQs a set of etypes, described by their relative properties, which are then used to build up the most suitable model for the ETG to be used as the schema of the final EG, which in Figure \ref{fig:pipeline} is called the \textit{ETG model}. The ETG model designed using the etypes and properties extracted from CQs can be then extended by extra etypes and properties suggested by the datasets. This extension is optional but suggested to allow for future expansions, as well as to increase the reuse of the available datasets. Notice that the availability of data would make this step low cost, in particular, if taken into account since the early stages. The \textit{ETG Modeling} activity aims to design the ETG model by shaping it as an EER diagram, in other words transforming the CQs into the ETG model. In parallel, the \textit{Dataset Selection} activity finalizes the selection of datasets, previously collected, by filtering out those that don't match the ETG model produced. 


The \textit{Knowledge Alignment} phase takes as input the ETG model previously generated, plus the selected set of datasets and reference ontologies. The main objective of this phase is to create the ETG for the final EG. The key observation is that the ETG is built to be as much as possible reusable, thus in turn enhancing the shareability of the whole EG. To this end, the input ETG model is itself a possible solution, nevertheless, it fits too much the CQs, by including definitions of etypes and properties less reusable for different purposes. To achieve the desired level of shareability, the more reused etypes and properties taken from the reference ontologies in inputs, are used to build the ETG (called ETG in Figure \ref{fig:pipeline}, without losing the purpose-specific semantic structure designed in the ETG model. This step is performed by the \textit{ETG Alignment} activity, implemented via the Machine Learning algorithm described in \cite{2020-KR}. The algorithm takes as input the set of reference ontologies and the ETG model while producing in output the final ETG. The resulting ETG is verified for compliance with the input datasets before the final approval. In parallel, the \textit{Dataset Cleaning} activity applies cleaning and formatting operations over the datasets in order to align data types and data formats with the ETG produced.
\\ \indent In the last phase, called \textit{Data Integration}, the inputs considered are the ETG and the datasets previously cleaned and formatted. The objective of this phase is to build the EG by integrating together the schema and data resources handled along the previous phases, according to the initial purpose. A single activity is present in this phase, called \textit{EG Generation}, which aims to merge the ETG and the datasets, by using a data mapping tool called \textit{KarmaLinker}, which consists of the \textit{Karma} data integration tool \cite{knoblock2015exploiting} extended to perform Natural Language Processing on short sentences (i.e., what we usually call \textit{the language of data}) \cite{2016-Bella1}. Using the tool the data values, in the datasets, are mapped over the etypes and properties of the ETG. In this step, the final EG's entities are generated and, whenever they are discovered to be different representations of the same real-world entity, merged into a single valid representation. These activities are fully supported by Karmalinker. The process described in this phase is iteratively executed for the list of datasets selected in the previous phase, processed sequentially. The process concludes with the export of the EG into an RDF file.
\\ \indent The key observation is that the desired middle-out convergence, as described in the introduction, has been implemented in two separate sub-processes, executed in parallel within each phase, one operating at the schema level, the other at the data level (blue and green boxes in Figure \ref{fig:pipeline}). During this process, the initial purpose keeps evolving building the bridge between CQs, datasets and reference ontologies. To enforce the convergence of this process, and also to avoid making costly mistakes, each phase ends with an evaluation activity (\textit{Eval} boxes in Figure \ref{fig:pipeline}).
The specifics of this activity are described in Section \ref{sec7_evaluation}. Here it is worth making two observations. The first is that this evaluation is driven by the non-functional requirements provided by the purpose. The second is that, within each phase, the evaluation aims to verify whether the target of that phase is met, namely: aligning CQs with datasets and ontologies in phase 1, thus maximizing the reuse of existing resources; aligning the ETG model with the datasets in phase 2, thus guaranteeing the success (purpose's specificity) of the project, and aligning ETG and ontologies in phase 3, thus maximizing EG sharability. The evaluation in phase 4 has the goal of checking that the final EG satisfies the requirements specified by the purpose. As from  Figure \ref{fig:pipeline}, if the evaluation results don't satisfy acceptable thresholds, in any of the steps in which is expected, the process goes back to the evaluation step of the previous phase. In the extreme case of a major early design mistake, it is possible to go back from phase 4 to phase 1. 

\section{Data reuse in \itelos} 
\label{sec5_reuse}

During the Inception and Modeling phases, iTelos aims at enhancing the reuse of existing resources in order to reduce the effort in building the EG. The main goal of the first two phases, in this perspective, is to transform the specifications provided by the input purpose into the ETG model. The process achieves this objective by following five different steps, along the two phases:

\begin{enumerate}
    \item initial formalization of the purpose into a list of natural language sentences, each informally defining a CQ;
    \item extraction, from each CQ, of a list of relevant etypes, and corresponding properties, thus formalizing slightly more the initial purpose;
    \item selection of the datasets whose schema matches the etypes extracted from the CQs in the previous step;
    \item generation of the list of etypes and relative properties extracted from the dataset selected previously, by matching them with those extracted from the CQs;
    \item construction of the ETG model by using the etypes, and properties, from the previous step.
\end{enumerate}
 \noindent
Most of the work is done during the inception phase, which covers steps 1-4, while step 5 happens during the modeling phase, where the choices made during the previous phase are selected and exploited to build the ETG model. Nevertheless, if the modeling phase's evaluation doesn't produce the desired results, there is the opportunity of backtracking in order to fix wrong choices made during the inception phase. With the aim of enhancing the reusability, the key idea is to classify the three types of resources (CQs, ontologies, datasets) handled during the first two phases, into three categories defining how reusable such resources are. Moreover, along the execution of the two phases, the resources are handled through a series of three iterative executions each corresponding to a specific category, following a decreasing level of reusability. The categories are defined as follows (see \cite{KD-2022-Giunchiglia-Popularity} for a first description of this specific three-level categorization):
 


\vspace{0.2cm}
\noindent
\textit{Common}: the resources classified in this category are those used to express aspects that are common to all domains, even outside the purpose-specific domain of interest. Usually, these knowledge resources correspond to abstract etypes specified in \textit{upper-level ontologies} \cite{DOLCE}, e.g., \textit{person}, \textit{organization}, \textit{event}, \textit{location}, or even etypes from very common domains, usually needed in most applications, e.g., \textit{Space} and \textit{Time}. Etypes and properties classified as common by \itelos\, correspond to what in knowledge organization are called \textit{Common Isolates} \cite{srr67}. Moreover, \itelos\ classifies as common data resources those that are provided by Open Data sites.
   
\vspace{0.2cm}
\noindent 
\textit{Core}: the resources classified in the core category express the most core aspects in the purpose-specific domain of interest. The information carried by these resources is fundamental, given its relevance to the purpose, it would not be possible to develop EG without it. Consider for instance the following purpose:
    
\vspace{0.1cm}
\noindent
    "\textit{There is a need to support the health of European citizens by opening up new ways to access their health data where needed (independently of the specific country's healthcare system). To this end, interoperable health data needs to be produced, by integrating local data from different countries, thus represented through different medical standards and languages.}"\footnote{This example, as well as all the follow-up material, as described below in the paper, has been extracted from the InteropEHRate EU project.}

\vspace{0.1cm}
\noindent
In this example, core resources could be those data values reporting patient's health information, like medication details, drugs, medical codes (e.g., Health, interoperability standards of various types). Examples of common resources, instead, are general information about the patient, like name, surname and date of birth, as well as upper-level ontologies that can be found in the repositories mentioned above. It is important to notice that in general, data are harder to find than ontologies, in particular when they are about sensible sectors like healthcare, where personal data is strongly considered.
    

\vspace{0.2cm}
\noindent
\textit{Contextual}: the resources classified in this category carry, possibly unique, information of the purpose-specific domain of interest. While the core resources contribute towards having a meaningful application, the contextual ones create added value in the EG developed, by making explicit the difference with respect to the competitors. In the example above, the resources classified as contextual can be the translations of health data which need to be included in the output interoperable health data in order to be exploited in different countries. At the schema level, contextual etypes and properties are those which differentiate the ontologies which, while covering the same domain, actually present major differences. \cite{2017-ICCM} presents a detailed quantitative analysis of how to compare these ontologies. Data-level contextual resources are usually not trivial to find, given their specificity and intrinsic. In the various applications that we have developed in the past, this type of data has turned out to be a new set of resources that had to be generated on purpose for the application under development, in some cases while in production. In the above example, some contextual data resources are the mappings between different medical standards codes. They are not always available, with the direct consequence that the mappings have to be produced on purpose by hand.

The overall conclusive observation is that the availability, and thus the reusability, of resources, and of data in particular, decreases from common to contextual category, or in other words from more infrastructural data to more application-specific data. Moreover, as described in \cite{KD-2022-Giunchiglia-Popularity}, the decrease in reusability goes in parallel with the increase of pre-processing required to create and handle more contextual data.


\begin{table}[]
    \centering
    \caption{CQs categorized in the three reusability categories.}
    \begin{adjustbox}{max width=\textwidth}
    \begin{tabular}{|c|l|l|l|}
        \hline
        Number & Question & Action & Category \\
        \hline
        1 & \makecell[l]{Which is the patient's general information\\ reported in the Patient Summary?} & Return the interoperable Patient Summary & Common \\
        \hline
        2 & Which are the medical information for a patient's medication ? & \makecell[l]{Return the medication information, \\ as well as the code of the involved drugs} & Core \\
        \hline
        3 & Which is the international version of an Italian medication? & Return  multilingual interoperable medication information & Contextual\\
        \hline
    \end{tabular}
    
    \end{adjustbox}
    \label{tab:cq}
\end{table}
\noindent
 Let us see an example of how the five elaboration steps listed at the beginning of the section intermix with the three categories above.
 Table \ref{tab:cq} shows some CQs, extracted from the case study introduced above (step 1). Notice how, already in this step, CQs are categorized as being common, core or contextual (last column) and how this is done after transforming the text from the purpose (left column) into something much closer to a requirement for the EG (central column). Table  \ref{tab:cq-etypes} then reports the etypes and properties extracted from the CQs (step 2). While this is not reported in the table for lack of space, these etypes and properties inherit the category from the CQs. Notice that it is possible to have an etype that is core or common or contextual with properties in all three categories; this being a consequence of the fact that an etype can be mentioned in multiple (types of) CQs.
 
 \begin{table}[]
    \centering
    \caption{Etypes and properties from the CQs.}
    \begin{adjustbox}{max width=\textwidth}
    \begin{tabular}{|c|l|l|}
        \hline
        CQ Number & Etypes & Properties \\
        \hline
        1 & Patient, Vital\_signs, Care\_plan & \makecell[l]{Patient\_identifier, Name, Surname, Date\_of\_birth,\\ Blood\_pressure, Care\_plan\_category}\\
        \hline
        2 & Medication, Drug  & \makecell[l]{Medication\_subject, Medication\_date, Drug\_identifier,\\ Coding\_system, Code\_value}\\
        \hline
        3 & Medication, Translation & \makecell[l]{Target\_language, Source\_language,\\ Medication\_dosage\_instruction, , Medication\_text\_note, }\\
        \hline
    \end{tabular}
    \end{adjustbox}
    \label{tab:cq-etypes}
\end{table}
\noindent
The next step is to select the datasets (step 3). Let us assume that the dataset used to (partially) answer the queries in Table \ref{tab:cq} generates the properties  matching the CQs as in Table \ref{tab:data} (step 4). As an example of matching, compare the attribute \textit{CD-ATC} in Table \ref{tab:data} with the property \textit{Code\_value} of CQ n°2 in Table \ref{tab:cq-etypes}.
Notice how the ordering of the analysis (from common to core to contextual) creates dependencies that may drive the choice of one dataset over another. As an example,  \textit{Medication\_dosage\_instruction} might be needed as a contextual property, but to properly define it, we also need the core etype \textit{Medication}. Analysing the resources required following such an order over the three reusability categories implies a major usage of more reusable resources. Concluding the process in the first two \itelos\ phases, steps 5 builds the ETG model, reported in Figure \ref{fig:eer}, containing etypes and properties from the tables above. 

 \begin{table}[ht!]
    \centering
    \caption{Dataset's attributes classified according to the reusability categories.}

    \begin{adjustbox}{max width=\textwidth}
    \begin{tabular}{|c|c|c|c|}
        \hline
        Attributes & Description & Type & Category   \\
        \hline
        ID-PATIENT & identifier of a patient into the dataset & string & Common\\
        \hline
        firstname & name of a patient into the dataset & string & Common\\
        \hline
        familyname & surname of a patient into the dataset & string & Common\\
        \hline
        CD-ATC & drug's medical code specified in the dataset & string & Core\\
        \hline
        beginmoment/date & date of the medication specified in the dataset & date-time & Core\\
        \hline
        content/text & textual information about a medication  & string & Contextual\\
        \hline
    \end{tabular}
    \end{adjustbox}
    \label{tab:data}
\end{table}

\section{Data sharing in \itelos} 
\label{sec6_share}

In the knowledge alignment phase, the main objective is to enhance the shareability of the EG to be produced, by building an ETG that can be shared for different future purposes. The approach followed by \itelos\ in order to achieve such a shareability is to build the ETG by reusing as much as possible the etypes and properties coming from well-formed standard reference ontologies, but under the overall guidance of the ETG model previously built. While the ETG model keeps the generation of the ETG focused on the initial purpose, the exploitation of reference ontologies improve the possibility to share it in different domains, where such ontologies are already involved in. The key observation is that the alignment mainly concerns the common and, possibly, the core types with much smaller expectations on contextual etypes. Notice that, in retrospect, the alignment with the most suitable ontology can enable the reuse of the data produced. As an example, the selection of FHIR \footnote{\url{http://hl7.org/fhir/}} as the reference ontology for medical data, ensures the compliance of the ETG produced with a huge amount of healthcare information which is already structured using FHIR. This type of decision should be made during the inception phase; if discovered here, it might generate backtracking.


\itelos\ implements the construction of a shareable ETG by adapting the \textit{Entity Type Recognition (ETR)} process proposed in \cite{2020-KR}. This process happens in three steps. The first step is the
\textit{selection of ontologies}. This step aims at selecting the set of reference ontologies that best fit the ETG model. As from \cite{2020-KR}, this selection step occurs by measuring each reference ontology according to two metrics, which allow:
    \begin{itemize}
        \item to identify how many etypes of the reference ontologies are in common with those defined in the ETG model,  and
        \item to measure a property shareability value for each ontology etype, indicating how many properties are shared with the ETG model etypes. 
    \end{itemize}
\noindent The output of this first step is a set of selected ontologies that best cover the ETG model, and that have been verified to fit the dataset's schema, at both etypes and etype properties levels.
The second step is the
\textit{Entity Type Recognition}(ETR). The main goal of this step is to predict, for each etype of the ETG model, which etype of the previously selected ontologies, analyzed one at a time, best fits the ETG. In practice, the ETG model's etypes are used as labels of a classification task. Such execution, as mentioned in \cite{2020-KR}, exploits techniques that are very similar to those used in ontology matching (see, e.g., \cite{2012-Giunchiglia3}). The final result for this step is a vector of prediction values, returning a similarity score between the ETG model's etypes and the selected ontology etypes. 
The third step is the \textit{ETG generation}. This step identifies, by using the prediction vector produced in the previous step, those etypes and properties from the reference ontologies which will compose the final version of the ETG. It is important, in this activity, to preserve the mapping with the datasets' schemas; whenever this is problematic, this becomes a possible source of backtracking.

\begin{figure}[ht!]
    \centering
    \includegraphics[width=11cm,height=3cm]{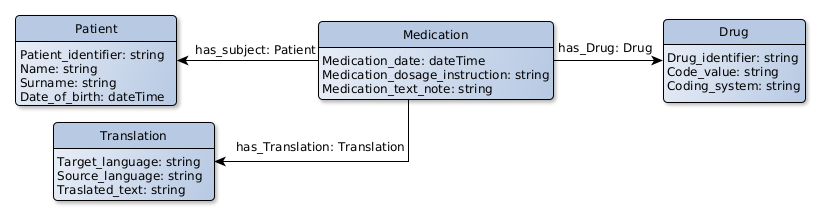}
    \caption{Portion of ETG model.}
    \label{fig:eer}
\end{figure}

\begin{figure}[ht!]
    \centering
    \includegraphics[width=9cm,height=3.5cm]{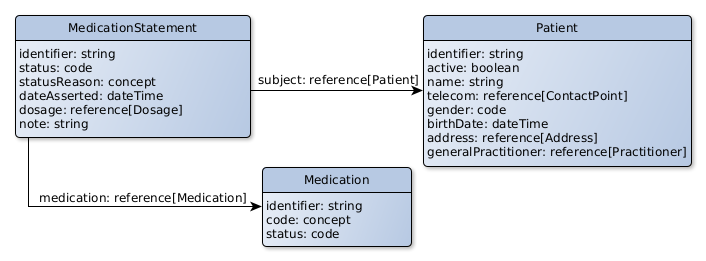}
    \caption{Portion of the \textit{FHIR} reference ontology.}
    \label{fig:fhir}
\end{figure}

\vspace{0.2cm}
\noindent
The distinction, among common, core and contextual etypes and properties, as well as their processing order, plays an important role in this phase and can be articulated as follows:
\begin{itemize}
    \item The common etypes should be adopted from the reference ontology, in percentage as close as possible to 100\%. This usually results in an enrichment of the top level of the ETG model by reusing existing representations of general cross-domain etypes (e.g., \textit{thing}, \textit{product}, \textit{event}, \textit{location}) that are usually less considered by developers due to their abstract nature, but which are fundamental for building an ETG, having all properties positioned in the right place, that can be shared among different domains and for different purposes. Moreover, this allows for a cross-domain alignment of those \textit{common isolates} (see Section \ref{sec5_reuse}) for which usually a lot of (open) data are publicly available (e.g., \textit{street});
    \item The core etypes are tentatively treated in the same way as common etypes, but the results highly depend on the reference ontologies available. Think for instance of the FHIR example above;
    \item Contextual etypes and, in particular, contextual properties are those most difficult to be found in existing reference ontologies. For this reason, such etypes and properties are mainly used for the high-level selection among the available ontologies, more than for the selection of single etypes inside those ontologies \cite{2020-KR}.
\end{itemize}
As an example of the process described above, compare the portion of ETG model as from Figure \ref{fig:eer}, with the portion of the \textit{FHIR} reference ontology in Figure \ref{fig:fhir}.\footnote{
 In the InteropEHRate project, reported as a use case in this paper, the matching between \textit{FHIR} and the ETG model has been done manually.} As it can be noticed, the \textit{FHIR} etypes \textit{MedicationStatement}, \textit{Medication} and \textit{Patient} can be matched to the ETG model etypes \textit{Medication}, \textit{Drug} and \textit{Patient}, respectively. As a consequence of such a matching, \itelos\ adopts the FHIR etypes and properties to compose the final version of the ETG. It is important to notice how the contextual etype, called \textit{Translation} in the ETG model, don't have any reusable counterpart in the FHIR reference ontology. For this reason, such a purpose-specific etype is taken from the ETG model to be part of the final ETG. 
 

\section{Alignment Evaluation} 
\label{sec7_evaluation}

\vspace{-0.1cm}

The evaluation activities implemented in the \itelos\ process (see Section \ref{sec4_process}) are based on a set of metrics applied to the intermediate outputs of the different phases, such as CQs, ETG and datasets, as well as to the final EG. For lack of space, below we describe only the application of three metrics to the specific phases in which they are exploited. In order to describe the evaluation metrics, let $\alpha$ and $\beta$ be two generic \textit{element} sets, where an element can be an etype or a property. We have the following:



\vspace{0.1cm}
\noindent
 \textit{Coverage}. This metric is used to measure the overlap between two element sets. In detail, how many etypes and properties can be found in both the sets are analyzed. Such an evaluation is computed as the ratio between the intersection of $\alpha$ and $\beta$ and the whole $\alpha$. 
 
 \vspace{-0.2cm}
    \begin{equation}
    \label{eq1}
    Cov = (\alpha \cap \beta) / \alpha
    \end{equation}
    
\vspace{-0.1cm}
\noindent For instance, during the inception phase (\textit{Eval(a)} in Figure \ref{fig:pipeline}), \(Cov\) plays a central role in evaluating the \emph{reusability} of potential datasets (via dataset schemas) with respect to the CQs. For each dataset, a high value of \(Cov\), both applied to etypes and properties, implies that the dataset is highly appropriate for the purpose. A low value of \(Cov\) implies minimal overlap between the purpose and the dataset, the consequence being non-consideration of the dataset for reuse, and possible modification of the (underspecified) CQs. 
    
\vspace{0.1cm}
\noindent
\textit{Extensiveness}. This metric quantifies the proportional amount of knowledge provided by any element set (such as $\beta$), in terms of sets of etypes or sets of properties, with respect to the entire knowledge considered (here $\alpha$ and $\beta$)
 \vspace{-0.1cm}
    \begin{equation}
    \label{eq2}
    Ext = (\beta - (\alpha \cap \beta)) / ((\alpha + \beta) - (\alpha \cap \beta))
    \end{equation}
\noindent During the Modeling phase (\textit{Eval(b)} in Figure \ref{fig:pipeline}), the evaluation utilizes \(Ext\) to measure how much the ETG model extends the set of CQs, with the objective of building the most suitable model for the purpose. To that end, a high value of \(Ext\) is evaluative of the fact that the ETG extends the scope of the CQs, by indicating a limited contribution of the CQs in generating the ETG model. On the other hand, low values of \(Ext\) are evaluative of the fact that CQs have contributed significantly towards the construction of the ETG model. 

\vspace{0.1cm}
\noindent
 \textit{Sparsity}. This metric quantifies the element-level difference between any number of similar element sets and is defined as the sum of the percentage of $\alpha$ not in $\beta$, and vice versa.
  \vspace{-0.2cm}
    \begin{equation}
    \label{eq3}
    Spr = ((\alpha + \beta) - 2(\alpha \cap \beta)) / ((\alpha + \beta) - (\alpha \cap \beta))
    \end{equation}
     
\vspace{-0.2cm}
\noindent 
In the knowledge alignment phase (\textit{Eval(c)} in Figure \ref{fig:pipeline}), our principal focus is to utilize \(Spr\) for ensuring the \emph{shareability} of the ETG. We incrementally enforce shareability by ensuring a required threshold of \(Spr\) between the ETG and each of the reference ontology. Such a threshold indicates that the ETG contains axioms reflective of \emph{contextual knowledge}. Nevertheless, this evaluation aims at maximizing the adoption of reference ontology's knowledge for common and core elements. 

\vspace{0.2cm}
\noindent
All the above metrics operate at the schema level. But they do not say anything about the results of integrating the datasets as caused by the semantic heterogeneity existing among them. Let's assume the following situation, in order to describe the data level evaluation criteria adopted by \itelos\ . Using \textit{KarmaLinker}, we are going to integrate a new dataset $D_1$ into a partially built EG. Moreover, $D_1$ has an etype $E_1$, with its property set $A_1$ and the EG already contains an etype $E_2$, with its property set $A_2$. Then we have the following possible situations:
\begin{itemize}
\vspace{-0.1cm}
    \item ($E_1 = E_2$) The etype $E_1$ in $D_1$ is already integrated in the EG. As a consequence of integrating $D_1$ into the EG, there is an increase in the number of entities represented through the etype $E_1$, and the EG, after the integration, will be enriched by the connections of the new entities. Nevertheless, this situation includes two different sub-cases: (i) When $A_1 = A_2$, i.e., the two etypes share the same set of properties, conflicts (different values for the same property) are possible between the value set of $A_1$ and $A_2$. In this case, \itelos\ aims at identifying how many of such conflicts appear during the integration, as well as the number of properties remaining without a value (null or insignificant value) as a result of the integration. (ii) In the second sub case, i.e., $A_1 \neq A_2$, the two etypes have different sets of properties. As a consequence, there are no conflicts between the value set of $A_1$ and $A_2$, and we obtain a greater integration over the entities of $E_1$.
    \item ($E_1 \neq E_2$) The etype $E_1$ in $D_1$ is not yet present in EG. The consequence is that by integrating $D_1$ into the EG, we are increasing the number of etypes (and of the entities of such an etype) in the final EG. Once again, in this situation, we can differentiate two sub-cases: (i) When $E_1$ and $E_2$ are linked by at least one object property, the resulting EG, after the integration of $D_1$, will be a connected graph. In this sub-case, \itelos\ aims at evaluating the level of connectivity of the graph by identifying how many entities of $E_1$ have not null values for the object properties linking $E_1$ with the rest of the EG's entities. In the second sub-case, there are no object properties linking $E_1$ and $E_2$ (or $E_1$ with any other etype in EG). As a consequence, the EG after the integration of $D_1$, will not be connected and the information carried by $D_1$ cannot be reached navigating the EG. Therefore, the integration of $D_1$ doesn't increase the EG's connectivity.
\end{itemize}


\vspace{-0.1cm}
\noindent
The data driven criteria briefly introduced above are crucial for the evaluation of the quality of the final EG. At the moment, these characteristics of the EG have been evaluated by considering the above criteria during the process as well as over the final EG. However, a more precise set of metrics for the evaluation of the quality of the final EG is under definition.

\section{The InteropEHRate case study} 
\label{sec8_case}
\noindent
The \itelos\ methodology has been applied to the InteropEHRate EU project. The objective of the project was to keep European citizens fully in control of their own medical data. To achieve this goal, the partners put their focus on two key aspects: 
\begin{enumerate}
    \item To support the citizens and healthcare practitioners through a cross-country digital health infrastructure, composed of mobile applications as well as hospital and third-party digital services \cite{kiourtis2022electronic}. Nevertheless, such an infrastructure is not sufficient to completely achieve the objective of the project.
    \item The second key aspect considered in the project was the production and exploitation of interoperable Electronic Health Records (iEHRs). An iEHR is a medical data resource that can be exploited by citizens and healthcare practitioners (as well as researchers), regardless of the European country they are in, even if the country is different from what they are used to living. In detail, IEHR aims at aligning the heterogeneity currently present in medical data of different countries (or even within a single country) into a single multilingual data format (adopting the FHIR medical standard) easily exploitable in Europe thanks to the infrastructure provided by the project.
\end{enumerate}

\noindent In the context described above, \itelos\ has been exploited for the integration of local medical data coming from different European countries. In detail, some test sets of local data provided by four different hospitals located in Italy, Belgium, Romania, and Greece, respectively, have been used as input for the methodology, together with their reference schemas. The main problem to be solved, dealing with such local data, was the high level of heterogeneity between the data provided. A concrete example was the differences between the data provided by the Italian hospital, expressed using the HL7-CDA standard \footnote{\url{http://www.hl7.org/implement/standards/product_brief.cfm?product_id=7}}, and the Belgian hospital's data expressed through a Belgian medical standard called \textit{SumEHR}\footnote{\url{https://www.ehealth.fgov.be/standards/kmehr/en/transactions/summarised-electronic-healthcare-record-v20}}. The CDA data are structured in XML format, while the SumEHR are instead JSON files using different attributes to express the same medical information. Another major difference, between the two kinds of data, concerns the medical coding system adopted by the different standards. While the CDA supports the LOINC \footnote{\url{https://loinc.org/}} codes, the SumEHR uses a local coding system not recognized outside Belgium. Moreover, as described in \cite{KD-2022-Bocca.a}, another issue to be considered, in such a context, is the natural language heterogeneity leading to the need to produce multilingual iEHRs. As the main input regarding the schema layer, the FHIR medical ontology has been adopted as the reference ontology for the construction of an ETG structuring a KG able to maintain iEHRs. \itelos\ has been implemented in the project, by a semi-automatic process of building a KG able to maintain an interoperable and multilingual version of the local information provided by the involved hospitals. The data, extracted from such a KG, form iEHRs satisfying the project's requirements.   

During the final phase of IEHR, the project's pilots have been executed to test, in real-case scenarios, if the project's results have been able to meet the initial requirements. From the partners involved in the pilots, we collected feedback, which helped us to understand some strengths and weak points regarding the methodology applied in the IEHR context. The descriptions of such feedback are reported below:\\
\begin{itemize}
    \item \textit{(Strength)} The methodology, thanks to its focus on the reuse of existing data, helped in discovering data heterogeneity, over the medical data, between different countries, as well as in finding solutions to make such data interoperable within Europe. During the project, some suggestions, about how to improve the FHIR standard for a better level of interoperability, have been submitted to the HL7 association.
    \item \textit{(Strength)} The mapping between local and international knowledge, and data, existing resources are onerous to bootstrap but lightweight to maintain. In other words from a methodological point of view, a good modeling phase is hard to be executed, but it leads to building KGs easy to maintain and evolve. This enforces the reuse and share approach adopted by \itelos\ in building KG. 
    \item (\textit{Weakness}) The input standards knowledge and the support of the tools, provided by \itelos, are necessary but not sufficient. In the healthcare domain, the need for precision and correctness in producing and converting data from one context to one another (local to international) always requires human supervision.
    \item (\textit{Weakness}) The required human intervention limits the scalability of systematic approaches for KGs building. To this end, the \itelos\ methodology needs to be improved in order to provide the highest level of (semi)automation with the least human intervention required.
\end{itemize}
\section{Conclusion} 
\label{sec9_conclusion}
In this paper we have introduced \itelos, a novel methodology for the creation of purpose-specific KGs, adopting a \textit{circular} development process. By this, we mean that the implicit goal of \itelos\ is to enable the development of KGs via the \textit{reuse} of already existing resources, thereby, reducing the effort in building new data (KG based) which can be, in turn, highly \textit{reused} by other applications in future. Further, we have also described how \itelos\ has been used in the context of the InteropEHRate EU project, with the objective of producing multilingual iEHRs.

%

\section*{Acknowledgements} 
The research described in this paper was supported by the \texttt{InteropEHRate} project, a project of the EC Horizon 2020 programme, grant number 826106. We thank all the people from the University of Trento who supported us in the execution of this project, in particular: Danish Asghar Cheema, Ronald Chenu Abente. The acronym \texttt{IEHR} from the \texttt{InteropEHRate} project, has been freely adapted in this paper as \texttt{iEHR} which stands for \textit{interoperable Electronic Health Records.}

\bibliographystyle{splncs04}
\bibliography{KnowDivePubs}

\end{document}